\documentclass[10pt,twocolumn,letterpaper]{article}

\usepackage[pagenumbers]{cvpr} 

\usepackage[accsupp]{axessibility}  
\usepackage{graphicx}
\usepackage{amsmath}
\usepackage{amssymb}
\usepackage{booktabs}
\usepackage{multirow}
\usepackage{tabularx}
\usepackage[table]{xcolor}

\usepackage[symbol]{footmisc}

\usepackage[pagebackref,breaklinks,colorlinks]{hyperref}

\usepackage[capitalize]{cleveref}
\crefname{section}{Sec.}{Secs.}
\Crefname{section}{Section}{Sections}
\Crefname{table}{Table}{Tables}
\crefname{table}{Tab.}{Tabs.}

\def\cvprPaperID{8397} 
\def\confName{CVPR}
\def\confYear{2023}

\newcolumntype{C}{>{\centering\arraybackslash}X}

\begin{document}

\title{Inverting the Imaging Process by Learning an Implicit Camera Model}

\author{Xin Huang$^{1}$\footnote[1]{}, Qi Zhang$^{2}$\footnote[2]{}, Ying Feng$^{2}$, Hongdong Li$^{3}$, Qing Wang$^1$\footnote[2]{} \vspace{6pt}\\
$^{1}$ School of Computer Science, Northwestern Polytechnical University, Xi'an 710072, China \\
$^{2}$ Tencent AI Lab \qquad $^{3}$ Australian National University\\
}

\twocolumn[{%
\renewcommand\twocolumn[1][]{#1}%
\maketitle

\begin{center}
    \centering
    \captionsetup{type=figure}
    \includegraphics[width=0.95\hsize]{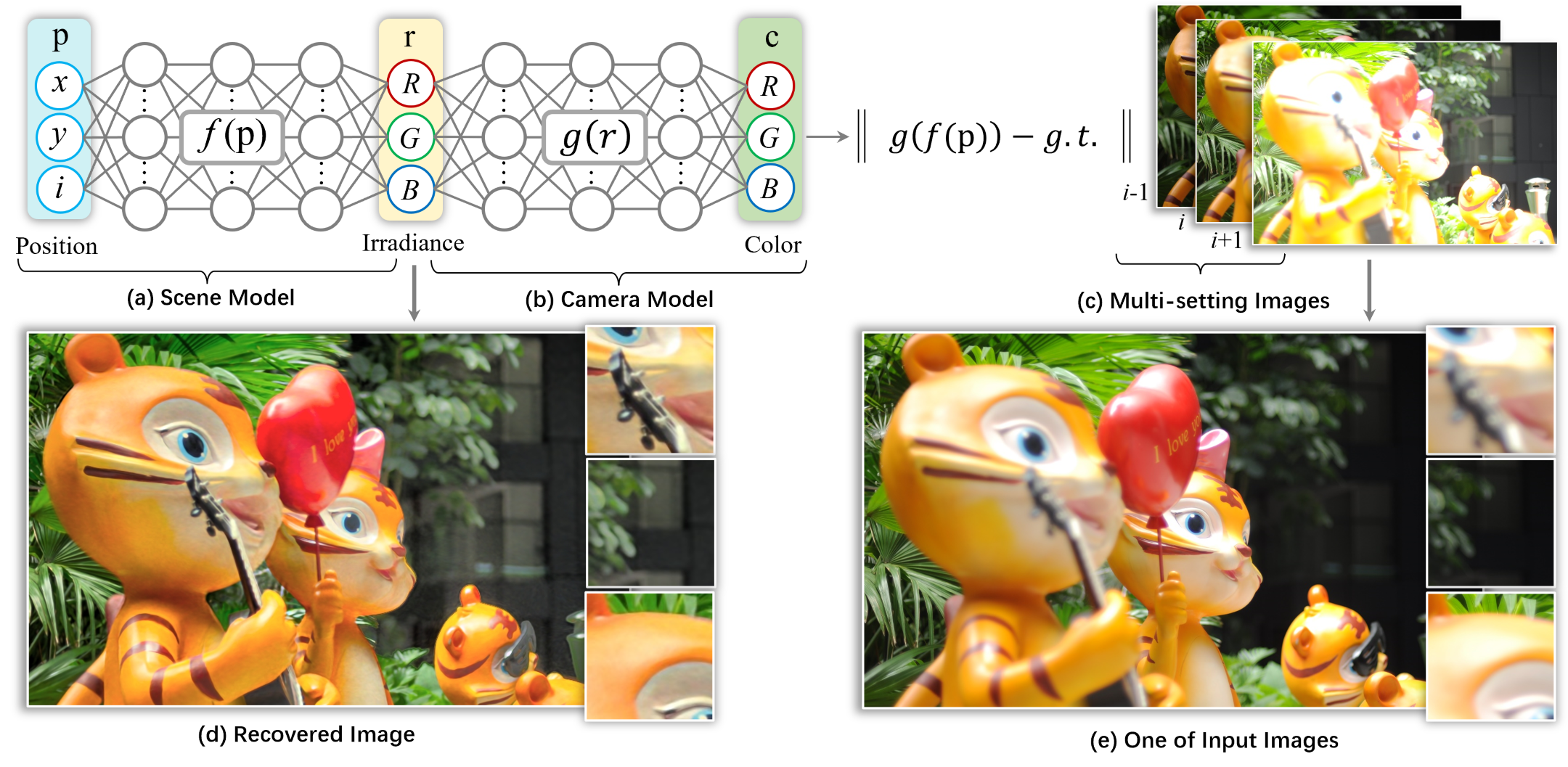}
    \vspace{-3mm}
    \caption{
    Our method solves inverse imaging tasks by learning an implicit neural camera model. The proposed framework consists of (a) a scene model representing scene contents and (b) a camera model simulating the camera imaging process. Given a pixel position $\mathbf{p}$ (2D pixel coordinate + 1D image index) at an image stack, the scene model maps it to corresponding irradiance value $\mathbf{r}$ (\textit{i.e.}, $\mathbf{r}=f(\mathbf{p})$), and then the camera model maps the irradiance $\mathbf{r}$ to a pixel intensity $\mathbf{c}$ (\textit{i.e.}, $\mathbf{c}=g(\mathbf{r})$). Two models are trained per scene and jointly optimized under the supervision of (c) a set of images captured with different camera settings (multi-focus and multi-exposure). After training, the scene irradiance have been implicitly encoded into the scene model (under an indirect supervision from the multi-setting images). We then remove the camera model and the scene model can render (d) all-in-focus HDR images by taking pixel positions as input.
    }  
    \label{fig:teaser}
\end{center}%
}]

\footnotetext[1]{Work was done during an internship at Tencent AI Lab.}
\footnotetext[2]{Corresponding authors.}

\begin{abstract}
Representing visual signals with implicit coordinate-based neural networks, as an effective replacement of the traditional discrete signal representation, has gained considerable popularity in computer vision and graphics. In contrast to existing implicit neural representations which focus on modelling the scene only, this paper proposes a novel implicit camera model which represents the physical imaging process of a camera as a deep neural network.  We demonstrate the power of this new implicit camera model on two inverse imaging tasks: i) generating all-in-focus photos, and ii) HDR imaging.  Specifically, we devise an implicit blur generator and an implicit tone mapper to model the aperture and exposure of the camera's imaging process, respectively. Our implicit camera model is jointly learned together with implicit scene models under multi-focus stack and multi-exposure bracket supervision.   We have demonstrated the effectiveness of our new model on a large number of test images and videos, producing accurate and visually appealing all-in-focus and high dynamic range images.  In principle, our new implicit neural camera model has the potential to benefit a wide array of other inverse imaging tasks. 
\end{abstract}

\vspace{-5mm}
\section{Introduction}
\label{sec:intro}

Using deep neural networks to learn an implicit representation of visual signal of a scene has received remarkable success  (\textit{e.g.}, NeRF \cite{mildenhall2020nerf}). It has been used to represent visual signals (\textit{e.g.}, images \cite{dupont2021coin,sitzmann2020implicit}, videos \cite{kasten2021layered,chen2021nerv}, and volume density \cite{mildenhall2020nerf}) with many impressive results.  Besides implicit scene  modelling (e.g., modelling scene radiance field via an MLP),  the physical imaging process of a camera is also important for the image formation process (\textit{i.e.}, from scene radiance field to RGB values of the sensor of a camera \cite{sturge1989imaging}). 

However, to the best of our knowledge, little in the literature has ever tapped into the issue of finding an implicit representation to model the physical imaging process of a camera.  Instead, most existing neural rendering methods assume that each pixel's RGB values are precisely the captured radiance field. In reality, before the light rays hit the imaging sensors, they need to pass through both the aperture and shutter, resulting in possible image blur caused by finite-sized aperture as well as varied dynamic range dictated by exposure time of the shutter. 

Moreover, the image signal processor (ISP) inside a digital camera may also alter the obtained image, e.g., luminance change, depth of field (DoF), as well as image noises.  
The above observation prompts us to address two questions in this paper:  
\begin{itemize}\setlength\itemsep{0pt}
    \item {\it Can we learn an implicit camera model to represent the imaging process and control camera parameters?} 
    \item {\it Can we invert the imaging process from inputs with varying camera settings and recover the raw scene content?}
\end{itemize}

Recently, learning-based methods simulating the mapping from raw images to sRGB images have been presented \cite{hu2018exposure, ouyang2021neural, yu2021reconfigisp}. They allow photo-realistic image generation controlled by the shutter or aperture, but inverse problems of raw image restoration are challenging to model.
Although a few NeRF-based methods have simply simulated cameras, they still face many issues, e.g., either RawNeRF \cite{mildenhall2021nerf} only models a camera forward mapping for controllable exposures or HDR-NeRF \cite{huang2021hdr} only builds a tone-mapper module with the NeRF on static scenes to inversely recover the high-dynamic-range (HDR) radiance. 
It is not clear whether a unified coordinate-based MLP module of different implicit camera models can be applied to various implicit neural scene representations for inverting the imaging process in a self-supervised manner, especially for dynamic scenes.

To this end, this paper proposes a novel implicit neural camera model as a general implicit neural representation.  Tested on two challenging tasks of inverse imaging, namely all-in-focus and HDR imaging, we have demonstrated the effectiveness of our new implicit neural camera model, as illustrated in Fig. \ref{fig:teaser}.  

The key contributions of this paper are: 
\begin{enumerate}\setlength\itemsep{0pt}
    \item We propose an interesting component, an implicit neural camera model including a \textit{blur generator} module (Sec. \ref{blur_generator}) for the point spread function and a \textit{tone mapper} module (Sec. \ref{tone-mapper}) for the camera response function, to model the camera  imaging process.
    \item We develop a self-supervised framework for image enhancement from visual signals with different focuses and exposures and introduce several regularization terms (Sec. \ref{sec:opt}) to encourage the modules of the implicit neural camera to learn corresponding physical imaging formulation.
    \item We showcase implicit image enhancement applications on images and videos fulfilled with the proposed framework, including forwardly controllable generation (changing exposures and focuses) or backwardly inverting restoration (all-in-focus and HDR imaging).
\end{enumerate}
 In the experiments, our method outperforms baseline methods in all-in-focus imaging and HDR imaging. Compared with traditional methods, our model can recover all-in-focus HDR images from fewer input images.

\section{Related work}
\noindent\textbf{Implicit Neural Representation}. 
Coordinate-based MLPs have been widely spread to represent a variety of visual signals, including images \cite{dupont2021coin,sitzmann2020implicit}, videos \cite{kasten2021layered,chen2021nerv} and 3D scenes \cite{mildenhall2020nerf}. Dupont \textit{et al}. \cite{dupont2021coin} demonstrate the feasibility of using implicit neural representation for image compression tasks.  Kasten \textit{et al}. \cite{kasten2021layered} introduce a coordinate-MLP-based framework that decomposes and maps a video into a set of layered 2D atlases, which enables consistent video editing. 
However, these methods only focus on the representation of the visual signal and ignore the camera model which is also an important component of the whole implicit representation. Neural radiance fields (NeRF) representation models a radiance field with the weights of a neural network, which can render realistic novel views. \cite{mildenhall2020nerf,yu2021pixelnerf,martin2021nerf,barron2021mip,boss2021nerd,li2021neural,ma2021deblur,chen2021hallucinated}. Most of the NeRF methods assume input images are of a consistent camera setting. However, without modeling the camera, it's difficult for them to handle the input with varying camera settings (modern cameras always adjust the exposure and focus automatically). Most recently, some NeRF-based methods \cite{kaneko2022ar,wu2022dof,huang2021hdr} focus on modifying the defocus blur or exposures of novel views. However, it's hard for them to control both exposures and defocus blur simultaneously. Particularly, Huang \textit{et al}.  \cite{huang2021hdr} learn the global tone-mapping process from radiance to image intensity, which enables them to reconstruct the HDR radiance field. However, they only model tone-mapping with NeRF on static scenes. It's challenging for them to deal with the dynamic scenes and the inputs with other varying camera settings.

\noindent\textbf{HDR Imaging}.
High Dynamic Range (HDR) imaging is a technique that recovers images with a superior dynamic range of luminosity. Debevec and Malik \cite{debevec2008recovering} propose the classic method for HDR imaging. They capture a set of images with different exposures and then merge those LDR images into an HDR image by calibrating the camera response function (CRF). However, they may cause ghost artifacts when the images are captured by hand-held cameras or on dynamic scenes. To overcome this, some two-stage approaches have been developed \cite{grosch2006fast,jacobs2008automatic,tursun2015state,kalantari2017deep,yan2019robust}. They first detect and remove the motion regions in the input images, and then merge the processed images into an HDR image.  Recently, several methods that do not require optical flow are proposed for HDR imaging of dynamic scenes \cite{wu2018deep,yan2019attention, yan2020deep, niu2021hdr}. They formulate the HDR imaging as an image translation problem from the input LDR images to the HDR images. However, the learning-based HDR imaging methods always need the HDR image as supervision. In contrast, our method is trained per scene in a self-supervised manner only requiring the input LDR images.

\noindent\textbf{Multi-focus Image Fusion}. Multi-focus Image Fusion (MFIF) has been studied for over 30 years \cite{zhang2021deep}, and various algorithms have been proposed. Li \textit{et al}. \cite{li2013image} propose a matting-based method to fuse the focus information from input images. Liu \textit{et al}. \cite{liu2017multi} propose the first CNN-based supervised MFIF method, which learns a decision map for the fusion of two source images. Inspired by this work, several works \cite{tang2018pixel,wang2020novel,yang2019multilevel} have been conducted to improve the prediction of decision maps. While other methods \cite{zhang2020ifcnn,li2019multi} directly map the source images to fused images via an encoder-decoder architecture. Moreover, GANs have also been applied to MFIF. Guo \textit{et al}. \cite{guo2019fusegan} formulate the MFIF as an images-to-image translation problem and utilize the least square GAN objective to improve their method. Supervised methods require a large amount of training data with ground truth, but the all-in-focus images are hard to access, thus some unsupervised MFIF \cite{xu2020fusiondn, xu2020u2fusion} methods have been proposed. Recently, the first GAN-based unsupervised method, MFF-GAN, is proposed by Zhang \textit{et al}. \cite{zhang2021mff}. An adaptive decision block is introduced to evaluate the fusion weight of each pixel based on the repeated blur principle. However, the MFIF methods produce all-in-focus images by fusing the input images, which makes them struggle with the unaligned input images or video frames. Wang \textit{et al}. \cite{wang2021deep} first propose a deep learning autofocus pipeline that can control the focus and generate all-in-focus images. However, they struggle to dynamic scenes with large camera motions and dynamic objects.

\begin{figure*}[t]
  \centering
  \includegraphics[width=\textwidth]{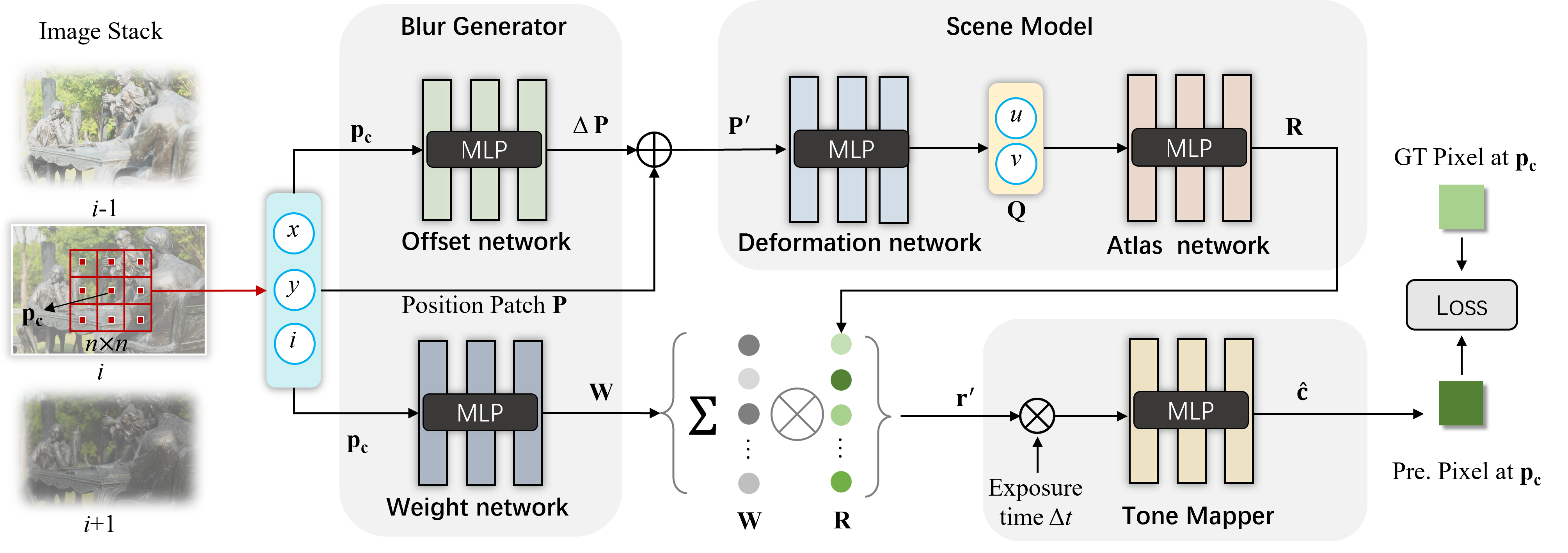}
  \vspace{-8mm}
  \caption{Illustration of our pipeline. $n\times n$ pixel positions $\mathbf{P}$ centered at $\mathbf{p}_c$ in the image stack or video sequence are fed into our model. Taking the $\mathbf{p}_c$ as input, our offset network and weight network outputs an offset patch $\Delta \mathbf{P}$ and a weight patch $\mathbf{W}$ respectively. The new position patch $\mathbf{P}'=\mathbf{P}+\Delta \mathbf{P}$ is then fed into the deformation network which predicts the corresponding 2D coordinate patch $\mathbf{Q}$ in the irradiance atlas. The atlas network maps $\mathbf{Q}$ into irradiance patch $\mathbf{R}$. We compute the blurry irradiance $\mathbf{r}'$ at position $\mathbf{p}_c$ by taking the sum of element-wise multiplication of irradiance patch $\mathbf{R}$ and weight patch $\mathbf{W}$. The blurry irradiance $\mathbf{r}'$ and exposure time $\Delta t$ are then mapped into pixel intensity $\mathbf{\hat{c}}$ by a \textit{tone mapper} which contains three small MLPs for R, G and B channels. During the inference, we remove the blur generator and tone-mapper and render all-in-focus and HDR images via the deformation network and atlas network. }
  \label{fig:pipeline}
  \vspace{-3mm}
\end{figure*}

\section{Method}
In a nutshell, the goal of our method is to invert the imaging process (\textit{e.g.}, all-in-focus imaging and HDR imaging) by learning an implicit camera model. The proposed framework is visualized in Fig. \ref{fig:pipeline}. Note that our method is trained per scene, which is similar to NeRF \cite{mildenhall2020nerf}. The input of our method is a set of coordinates (2D pixel coordinate + 1D image index) that denote the pixel locations at an image stack (multi-focus and multi-exposure images). Our model maps these coordinates to their corresponding pixel colors and minimizes the mean squared error between predicted colors and ground truth pixel colors for optimization. After training on the image stack, sharp scene irradiance is encoded in our scene model through indirect supervision from training images. This process resembles self-supervision since ground truth irradiance isn’t used for training. During inference, we remove the blur generator and tone-mapper module and render all-in-focus and HDR images by feeding pixel positions into the scene model.

\subsection{Neural Scene Representation}
 Inspired by the neural video representation method \cite{kasten2021layered}, we represent scene irradiance with two components: (1) a 2D atlas that records unique scene irradiance and (2) a deformation that matches each 3D pixel coordinate to its corresponding 2D point in the atlas. This deformation resembles the deformed field widely used in neural rendering for dynamic scenes\cite{pumarola2021d}. Specifically, we use an MLP-based deformation network $\mathcal{D}$ to map each 3D pixel position to a 2D coordinate in the atlas. Given a 3D pixel position $\mathbf{p}=(x, y, i)^\top$, where $(x,y)^\top$ is the pixel coordinate and $i$ is the image index, the deformation is given by: 
\begin{equation} \small
	\mathbf{q} = \mathcal{D}(\mathbf{p}),
	\label{eq:deformation} 
\end{equation} 
where $\mathbf{q}=(u,v)^\top$ denotes the 2D coordinate in the atlas.

Similarly, we use an atlas network $\mathcal{A}$ to represent the 2D atlas. Given the 2D coordinate  $\mathbf{q}$, the atlas network $\mathcal{A}$ maps it to the irradiance value at position $\mathbf{q}$. 
The process is formulated as: 
\begin{equation} \small
	\mathbf{r} \!=\! \mathcal{A}(\gamma(\mathbf{q})),
	\label{eq:atlas}
\end{equation}
where $\mathbf{r}$ denotes the irradiance and $\gamma$ denotes the positional encoding \cite{mildenhall2020nerf}.

\subsection{Blur Generator} \label{blur_generator}
Most classic image deblurring methods model the blur with a 2D convolution between image intensity and a Point Spread Function (PSF) \cite{campisi2017blind} that indicates the degree of blurring in an image. However, according to the physical imaging pipeline, the blur convolution should be applied to the irradiance rather than the image intensity \cite{chen2012theoretical}. We therefore extend the model to the linear irradiance domain. In general, the spatially invariant blur is mathematically formulated as:
\begin{equation}
    \mathbf{r}'=\mathbf{R}*\mathbf{W},
	\label{eq:blur}
\end{equation}
where $\mathbf{r}'$ represents the blurry irradiance, $\mathbf{R} \in \mathbb{R}^{n\times n}$ is the sharp irradiance patch, and $\mathbf{W} \in \mathbb{R}^{n\times n}$ is the PSF of the patch which is centered at $\mathbf{p}_c$. The operator $*$ denotes the 2D convolution. 

\begin{figure}[t]
  \centering
  \includegraphics[width=\linewidth]{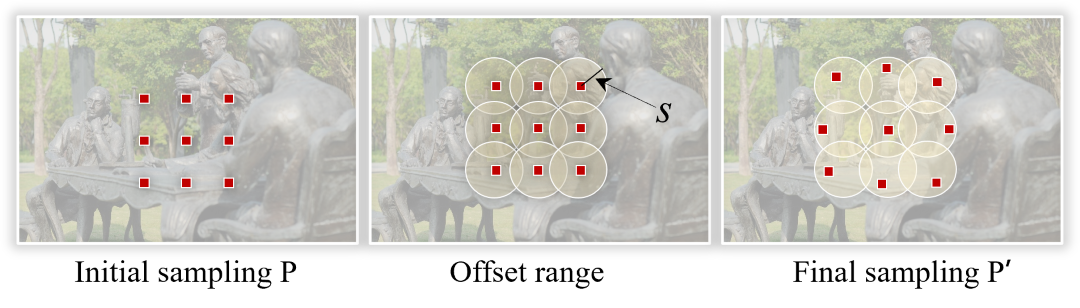}
  \vspace{-7mm}
  \caption{Sampling strategy of our method. For the sampling patch $\mathbf{P}$, 2D offsets for each position are learned to determine the final sampling patch $\mathbf{P}'$, which can flexibly modify the receptive field of the sampling. To limit the offsets, we set the maximum offset range (a circle with radius $s$), where $s$ is a hyperparameter.
  }
  \label{fig:samples}
  \vspace{-3mm}
\end{figure}

To render the blurry irradiance of point $\mathbf{p}_c$,  we once feed $n \times n$ (typically $3 \times 3$) positions into the scene model to predict the irradiance patch. However, a larger degree of blur always corresponds to a larger receptive field of the PSF and the $3 \times 3$ patch is too small to generate a large blur. One simple way to improve the receptive field of our sampling patches is using larger patches such as 5$\times$5 and $7\times7$. Unfortunately, it requires a large amount of computation and memory consumption. To solve this problem, we propose a new sampling strategy in which the receptive field of each sampled patch is flexible and optimized by a network. As shown in Fig. \ref{fig:samples}, given the center position $\mathbf{p}_c=(x,y,i)$ of the initial sampling patch $\mathbf{P}$, we use a lightweight offset network $\mathcal{O}$ to predict an offset patch $\Delta \mathbf{P}=(\Delta x,\Delta y,i)$. Therefore, the final sampling positions are set to $\mathbf{P} + \Delta \mathbf{P}$.
\begin{equation}
    \begin{split}
	\Delta\mathbf{P} &= \mathcal{O}(\mathbf{p}_c), \\
	\mathbf{P}' &= \mathbf{P} + \Delta \mathbf{P}.
	\label{eq:offset}
    \end{split}
\end{equation}
With the learned offsets, the network can automatically modify the receptive field of the sampling to match the large blur and only a little additional computational cost is introduced. Next, We feed the coordinates of sampling patch $\mathbf{P}'$ into the scene model, thus we can obtain the irradiance patch $\mathbf{R}$.

 The PSF depends on a set of factors, such as aperture size, focal length, object depth, etc. It's complicated to consider all these factors, especially the depth which is difficult to obtain. To simplify the model, each weight patch is optimized independently according to its 2D center coordinate $(x,y)$ and the image index $i$, where the index $i$ is used to embed the unique blur pattern of each training image. Thus, for an irradiance patch centered at position $\mathbf{p}_c=(x,y,i)$, we feed the $\mathbf{p}_c$ into a weight network $\mathcal{W}$ to predict the blending weight patch $\rm{W}$. The process is expressed as:
\begin{equation}
    \rm{W} =\mathcal{W}(\mathbf{p}_c).
	\label{eq:blur_weight}
\end{equation}

We further formulate the blur convolution as the sum of element-wise multiplication of weight patch and irradiance patch. Eq.(\ref{eq:blur}) thus is rewritten as:
\begin{equation}
    \mathbf{r}'=\sum_{\mathbf{x}\in{\mathbf{P}'}}\mathbf{r}(\mathbf{x})w(\mathbf{x}),
	\label{eq:blur_element}
\end{equation}
where $\mathbf{P}'$ is the final $n \times n$ sampled position. $\mathbf{r}(\mathbf{x})$ and $w(\mathbf{x})$ denote the irradiance and weight of position $\mathbf{x}$, respectively. 

\subsection{Tone Mapper} \label{tone-mapper}
To simplify the global tone-mapping process, we take the ISO gain and aperture size as implicit factors. Consequently, the tone-mapping function $f$ (also called CRF, Camera Response Function) is defined as:
\begin{equation}
    \mathbf{c}=f(\mathbf{r}\Delta t),
	\label{eq:tonemap}
\end{equation}
where $\mathbf{r}$ is the irradiance captured by a camera, $\mathbf{c}$ denotes the pixel intensity, and $\Delta t$ denotes the exposure time (shutter speed). We assume the exposure times are known because these can be obtained from the camera EXIF tags. Even if the exposure time is unavailable, we can jointly optimize the exposure time by taking it as a latent code. We also assume that the lighting change is insignificant and can be ignored. 

Similar to HDR-NeRF \cite{huang2021hdr}, we transform the global tone-mapping into the logarithm irradiance domain for better optimization. Specifically, we take logarithms to both sides of Eq. \eqref{eq:tonemap} (base $2$ is convenient, as we usually measure the exposure with exposure values (EVs)). Consequently, Eq. \eqref{eq:tonemap} is rewritten as:
\begin{equation}
    \log f^{-1}\left(\mathbf{c}\right) = \log \mathbf{r}+ \log \Delta t,
	\label{eq:tonemap_log}
\end{equation}
We further use a tone-mapping network $\mathcal{T}$ to implicitly represent $(\log{{f^{-1}}})^{-1} $. According Eq. \eqref{eq:blur_element} and Eq. \eqref{eq:tonemap_log}, our global tone-mapping is defined as:
\begin{equation}
    \mathbf{\hat{c}} = \mathcal{T} \left( \log \mathbf{r}' + \log \Delta t \right).
	\label{eq:tonemapper}
\end{equation}
where $\mathbf{r}'$ denotes the blurry irradiance (see Eq. \eqref{eq:blur_element}) and $\mathbf{\hat{c}}$ denotes the predicted blurry color. Generally, each color is consisted of red, green, and blue channels, so three small MLPs are used to model the \textit{tone mapper}.

\begin{figure*}[!t]
  \centering
  \includegraphics[width=\textwidth]{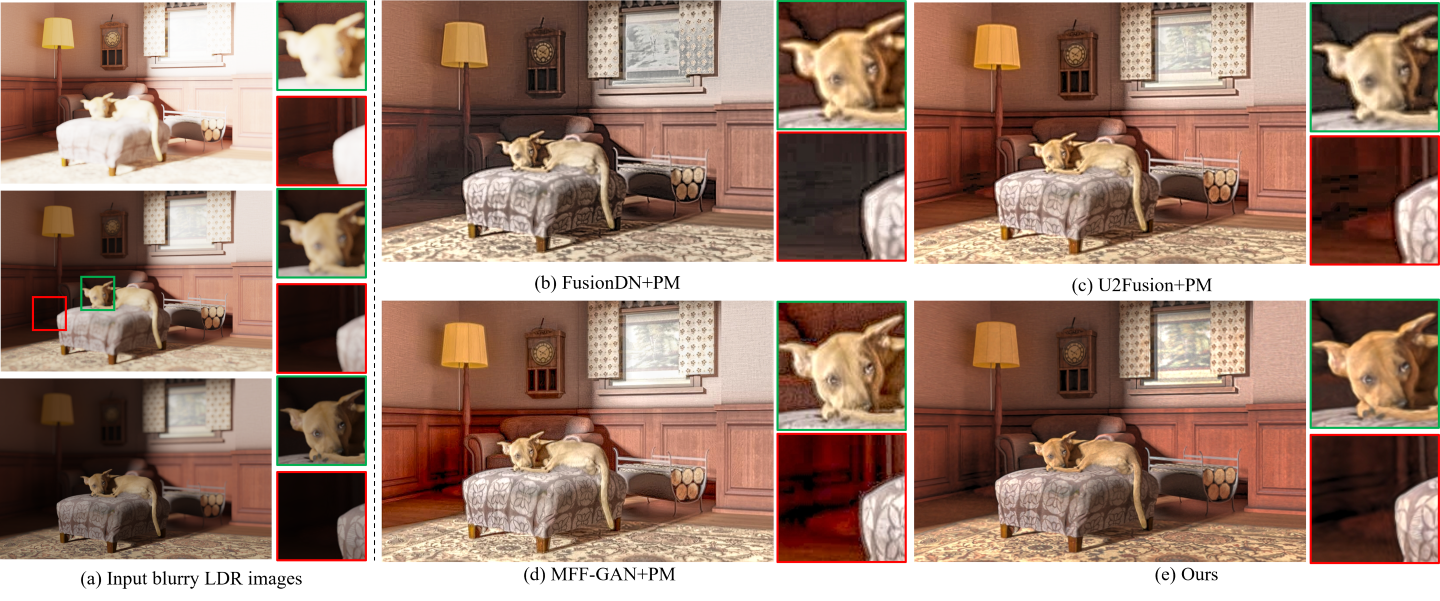}
  \vspace{-8mm}
  \caption{Example results of our method compared with two-stage methods on the MFME synthetic dataset. ``PM'' denotes the HDR imaging method in Photomatix \cite{photomatix} (a) Our input images with different focuses and exposures. (b-e) All-in-focus and HDR images produced by three two-stage methods and our method. The red and green insets show the zoom-in views of the images. All HDR images are tone-mapped for display.}
  \label{fig:deblur_HDR_syn}
  \vspace{-0.3cm}
\end{figure*}

\subsection{Optimization} \label{sec:opt}
To adapt to various scenarios, we design several loss terms for our neural camera, including color reconstruction loss, flow loss, white balance loss and gradient loss, to encourage our implicit neural camera to learn the imaging process correctly.

\noindent\textbf{Color reconstruction loss.} The color reconstruction loss is the main loss in our model. The predicted blurry LDR color is supervised by the input ground truth color. We minimize their mean squared error for optimization. Formally, the loss is given by:
\begin{equation}
	\mathcal{L}_{c} \!=\! \| \mathbf{\hat{c}} - \mathbf{c} \|^2_2,
	\label{eq:loss_color}
\end{equation}
where $\mathbf{\hat{c}}$ is our predicted color, and $\mathbf{c}$ is the ground truth color. 

\noindent\textbf{Flow loss.} Ideally, for each point in the scene, its corresponding pixels that are captured under different camera settings should be mapped consistently into the same position in the atlas. We find the deformation network can fulfill this expectation in some special cases, for example when input images are aligned or the motions of the camera and objects are small. To improve our results on challenging scenes, we use an off-the-shelf optical flow estimation method \cite{teed2020raft} to predict the optical flow of each point and design an explicit constraint. Specifically, our flow loss is defined as:
\begin{equation}
	\mathcal{L}_{f} = \| \mathbf{q_p} - \mathbf{q_{p^{*}}} \|^2_2,
	\label{eq:loss_flow}
\end{equation}
where $\mathbf{q_p}$ is the coordinate of position $\mathbf{p}$ in the irradiance atlas (Eq. \eqref{eq:deformation}), $\mathbf{p}^{*}$ is the corresponding position of $\mathbf{p}$ in the adjacent image. The estimated optical flow is not
always accurate due to the different focuses and exposures of input images. Therefore, during the training phase, the flow loss weight gradually decays to 0 over the course of optimization.

\noindent\textbf{White balance loss.} Theoretically, the irradiance recovered from multi-exposure images is relative to the true value with an unknown scale factor. The white balance of the irradiance is changed with the factor of each channel, thus the unconstrained scale factors estimated by our network lead to a random white balance in recovered irradiance. To tackle this problem, we introduce a loss to regularize the white balance. Formally, the white balance loss is given by:
\begin{equation}
	\mathcal{L}_{w} = \|\mathcal{T}(0) - \mathbf{c}_0 \|_2^2,
	\label{eq:loss_wb}
\end{equation}
where $\mathbf{c}_0$ is a hyperparameter which is generally set as the midway of the color value such as $0.5$ \cite{huang2021hdr}. The white balance loss encourages the unit irradiance to be mapped into $\mathbf{c}_0$, which allows us to regularize the white balance of the recovered irradiance.

\noindent\textbf{Gradient loss.}  CRF is a monotonically non-decreasing function \cite{debevec2008recovering}. Therefore, we add an explicit regularization to ensure the gradient of each point at the learned CRF is non-negative. We define the gradient loss as:
\begin{equation}
    \mathcal{L}_g = \rm{ReLU}(-\frac{\mathrm{d}\,\mathcal{T}(\mathbf{r})}{\mathrm{d}\mathbf{r}}),
    \label{eq:loss_gradient}
\end{equation}
where $\mathbf{r}$ is the input irradiance of \textit{tone mapper} $\mathcal{T}$ and $\rm{ReLU}$ denotes the ReLU (rectified linear unit) activation function.

\noindent\textbf{Total loss.} Finally, the total loss function is the weighted combination of the loss terms from \cref{eq:loss_color,eq:loss_flow,eq:loss_wb,eq:loss_gradient}:
\begin{equation}
	\mathcal{L}_{total} = \lambda_c \mathcal{L}_c + \lambda_f \mathcal{L}_f + \lambda_w \mathcal{L}_w + \lambda_g \mathcal{L}_g,
	\label{eq:loss_total}
\end{equation}
where $\lambda_c,\lambda_f,\lambda_w,\lambda_g$ are the weights of our loss terms. 

\section{Experiments}
\subsection{Implementation Details} \label{sec:details}
We employ positional encoding in atlas network $\mathcal{A}$, with the number of frequencies $7$. In \textit{blur generator} module, the size $n$ of the sampling path is set to $3$, and the maximum offset $s$ is set to $5$ pixels. The weights of loss terms are empirically set as $\lambda_c = 1$, $\lambda_f=100$,  $\lambda_w=1$ and $\lambda_g=100$. 
We use Adam optimizer with a learning rate of $1\times10^{-4}$ over the course of optimization. In each iteration, the batch size of point positions is set to $30,000$, and each model is optimized for around $150,000$ iterations. All experiments are conducted on a single V100 GPU. We train our model on an image stack, which takes about $1$ hours to finish. When training on video sequences, it takes about $5\sim 8$ hours to finish according the number of frames. 

\begin{table}[!t]
\small
\centering
\caption{Quantitative comparisons with two-stage methods on MFME synthetic dataset. Metrics are averaged over the synthetic scenes. PSNR-$\mu$, SSIM-$\mu$ and LPIPS are computed in the global tone-mapping domain. PSNR-$L$, SSIM-$L$ and HDR-VDP-2 are computed in the HDR domain. ``PM'' denotes the HDR imaging method in Photomatix \cite{photomatix}. }
\vspace{-0.2cm}
\begin{tabularx}{\linewidth}{@{}lCCCC}
\toprule
 & FusionD
 +PM & U2Fusio
 +PM & MFFGAN
 +PM & Ours\\
\hline
PSNR-$\mu$ & 17.39 & 29.19 & 27.88 & \textbf{31.25} \\
SSIM-$\mu$& 0.596 & 0.893 & 0.879 & \textbf{0.895} \\
LPIPS & 0.310 & 0.132 & 0.116 & \textbf{0.104} \\
\midrule
PSNR-$L$& 28.24 & 35.51 & 36.43 & \textbf{37.79} \\
SSIM-$L$& 0.697 & 0.960 & 0.953 & \textbf{0.963} \\
HDR-VDP-2& 45.40 & 55.27 & 54.47 & \textbf{58.11} \\
\bottomrule
\end{tabularx}
\vspace{-0.2cm}
\label{tb:deblur_HDR_syn}
\end{table}

\subsection{Datasets and Metrics}
\paragraph{\textbf{Datasets}.} We evaluate our method on three datasets: a multi-focus and multi-exposure (MFME) dataset, a multi-focus (MF) dataset, and a multi-exposure (ME) dataset. The MFME dataset consists of 4 real-world scenes and 4 synthetic scenes. Each scene contains 9 images of 3 different focuses and 3 different exposures. The real-world images are captured by a digital camera with a tripod, and the synthetic images are rendered in Blender \cite{blender}. We also render all-in-focus HDR images for the synthetic scenes, which allows for evaluating our method quantitatively. The MF dataset contains 8 real-world scenes. Two images focusing on the foreground and background respectively are captured for each scene. The resolution of the above images is $600\times 900$ pixels. The ME dataset contains 5 real-world dynamic scenes from the HDR imaging dataset \cite{sen2012robust}. Three images with different exposures are captured for each scene.  Note that the multi-focus images in the MFME dataset and MF dataset are unaligned since these images are captured by modifying the distance between the camera's lens and the image sensor.

\begin{figure}[!tb]
    \centering
    \includegraphics[width=\linewidth]{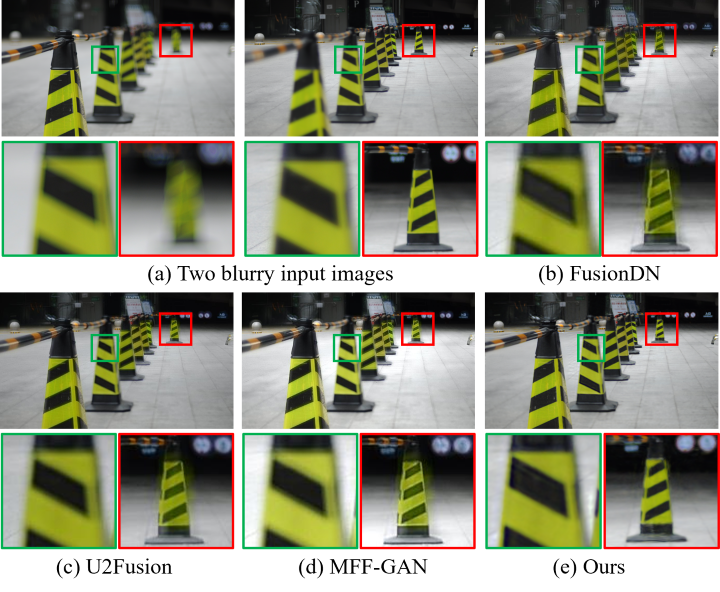}
    \vspace{-7mm}
    \caption{Example results of our method compared with MFIF methods on the MF dataset. (a) Two input images. One is near-focused and the other is far-focused. (b-e) All-in-focus results by MFIF methods and our method. The red and green insets show the zoom-in views of the images, \bf{better viewed on screen with zoom in.}}
    \label{fig:deblur}
    \vspace{-5mm}
\end{figure}

\paragraph{\textbf{Metrics}.} PSNR (higher is better), SSIM (higher is better) and LPIPS (lower is better) \cite{zhang2018unreasonable} are utilized as measurements for evaluation. Specifically, all HDR images are tone-mapped by $\mu$-law with $\mu=5000$, which is a simple classic global tone-mapping operator wildly used for HDR image evaluation \cite{kalantari2017deep,yan2019attention,prabhakar2020towards}.  We compute PSNR-$\mu$, SSIM-$\mu$ and LPIPS between predicted all-in-focus HDR images and ground truth images in the global tone-mapping domain. We also compute PSNR-$L$ and SSIM-$L$ for comparison in the original HDR domain. Moreover, another visual metric named HDR-VDP-2 (higher is better) \cite{mantiuk2011hdr} is also computed, which is specifically designed for the evaluation of HDR images.

\subsection{Evaluation}
\paragraph{\textbf{Baselines}.} To validate our method, we compare it with several methods on all-in-focus image restoration and HDR imaging tasks: (1) For all-in-focus image restoration, three SOTA methods for MFIF, MFF-GAN \cite{zhang2021mff}, U2Fusion \cite{xu2020u2fusion} and FusionDN \cite{xu2020fusiondn}, are selected as comparison methods. (2) For HDR imaging, we compare our method with three currently SOTA ghost-free HDR imaging methods, including HDR-GAN \cite{niu2021hdr}, AHDRNet \cite{yan2019attention} and DeepHDR \cite{wu2018deep}. (3) For all-in-focus and HDR image generation, we design two-stage comparison methods by combining MFIF methods with HDR imaging methods. The all-in-focus images are firstly recovered images by the aforementioned MFIF methods from the multi-focus images with consistent exposure. Taking all-in-focus images with different exposures as input, the final HDR images are reconstructed using the HDR imaging method in Photomatix \cite{photomatix} since we find the results by Photomatix is better than the standard static HDR imaging method by Debevec and Malik \cite{debevec1997recovering}. 

\begin{figure}[!t]
  \centering
  \includegraphics[width=\linewidth]{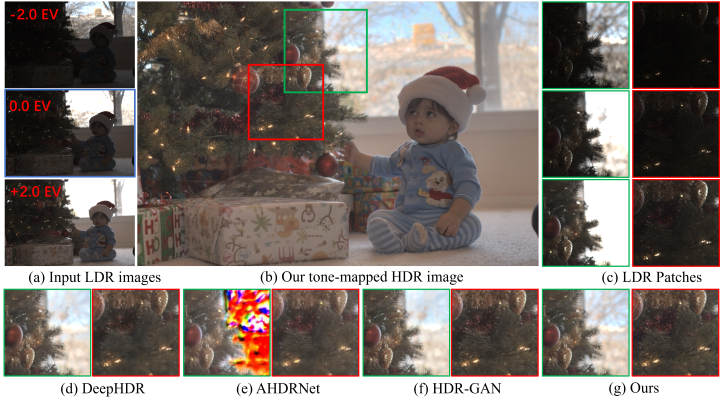}
  \vspace{-7mm}
  \caption{Example results of our method compared with HDR imaging methods on the ME dataset. (a) Three input images with different exposures. Exposure values (EVs) are shown in the upper left. The image highlighted with a blue box denotes the reference image. (b) Our recovered HDR image for the reference image. (c) zoom-in insets cropped from the input LDR images. (d-g) zoom-in insets cropped from the HDR images predicted by HDR imaging methods and our method. All HDR images are tone-mapped for display. }
  \label{fig:HDR}
  \vspace{-3mm}
\end{figure}

\noindent\textbf{All-in-focus and HDR imaging}. Figure \ref{fig:deblur_HDR_syn} presents qualitative comparisons with two-stage methods for all-in-focus HDR image restoration. Note that the two-stage methods require all 9 images as input, while our method are only trained 3 images of different exposures and focuses, as shown in Fig. \ref{fig:deblur_HDR_syn} (a). Compared to these two-stage methods, our method can recover image details from blurry LDR images (\textit{e.g.}, the dog in Fig. \ref{fig:deblur_HDR_syn}). All comparison methods produce ghosting near object boundaries (\textit{e.g.}, the boundary of the stool in Fig. \ref{fig:deblur_HDR_syn}), while our method produces sharp boundaries with a better visual experience. We also noticed that the results by FusionDN + PM show serious artifacts, likely due to color distortion in FusionDN's outputs from which Photomatix struggles to recover HDR images. Table \ref{tb:deblur_HDR_syn} shows quantitative comparisons on the MFME synthetic dataset. Our method outperforms state-of-the-art techniques on all metrics. Although U2Fusion + PM and our method have comparable SSIM values, our method performs better on object boundaries as discussed above. Results on LPIPS and HDR-VDP-2 metrics also validate that our method recovers all-in-focus HDR images with higher visual quality.

\noindent\textbf{Only all-in-focus}. Comparisons with the MFIF methods for all-in-focus image restoration are visualized in Fig. \ref{fig:deblur}. One can see that three MFIF methods produce ghost artifacts near the boundaries of objects. Generally, the focused and defocused boundary (FDB) is an important area where many algorithms do not perform well \cite{zhang2021deep}. In the patches near the FDB, both the focused area and the defocused area exist, which makes it difficult for these methods to produce plausible weights for image fusion. Compared with them, our method achieves sharp results without color distortion because our implicit camera learns the PSF rather than fusing the input images directly, which also indicates that our method has a better performance for recovering all-in-focus images from multi-focus images. 

\noindent\textbf{Only HDR imaging}. Comparisons with the HDR imaging methods for HDR image restoration are visualized in Fig. \ref{fig:HDR}. In this challenging scene, the sitting baby has small motions. DeepHDR struggles with complex textures and produces blurry results, as shown in the red insets. AHDRNet yields serious artifacts in over-exposed areas, and that is perhaps because AHDRNet is trained on the scenes without severe exposure deviation. HDR-GAN achieves acceptable results, but struggles to reconstruct over-exposed textures such as the twigs shown in the green insets. DeepHDR also has similar limitations. Compared with the above methods, our method recovers HDR results with clear textures and details in both under-exposed and over-exposed areas. Moreover, our method also achieves impressive results on moving objects, which can be referred from the face of the sitting baby, and that validates our method can deal with scenes with small motions.

\begin{figure}[!t]
  \centering
  \includegraphics[width=\linewidth]{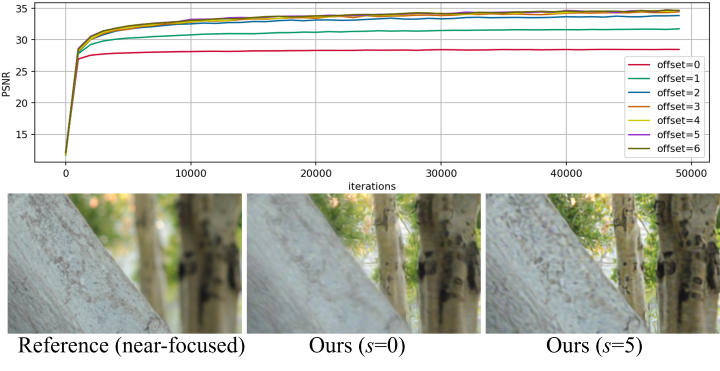}
  \vspace{-7mm}
  \caption{Results comparison of our method with different maximum offsets $s$ on an MF scene. Top row shows the PSNR during the training phase. Bottom row shows one of the input images and our all-in-focus results.}
  \label{fig:offsetsize}
  \vspace{-3mm}
\end{figure}

\subsection{Ablation Study} 
\paragraph{\textbf{Maximum offset.}} In our \textit{blur generator}, we introduce a maximum offset $s$ to control the receptive field of our sampling (see Sec. \ref{blur_generator}). To assess the impact of the maximum offset $s$, we train and test the framework with different $s$ values. The PSNR of the training phase is visualized in Fig. \ref{fig:offsetsize}. It's noticed that the PSNR improves with the value of maximum offset and the increase is gradually reduced, so we generally set $s=5$ in our experiments. Moreover, we can see that the recovered results with a larger offset ($s$=5) are sharper than the one with an offset set to $0$, which demonstrates the effectiveness of our sampling strategy.

\begin{figure}[t]
  \centering
  \includegraphics[width=\linewidth]{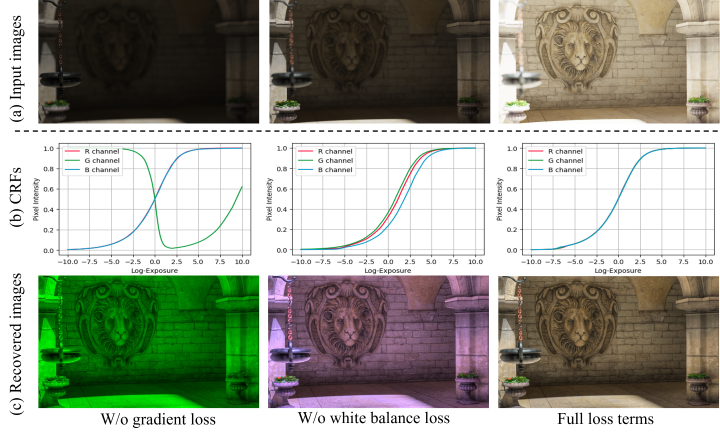}
  \vspace{-7mm}
  \caption{Results of our method with different loss terms on a synthetic MFME scene. (a) Three input blurry LDR images. (b) The learned CRFs by our \textit{tone mapper} module with different loss terms. (c) The all-in-focus HDR results by our method with different loss terms. \bf{Better viewed on screen with zoom in.}}
  \label{fig:loss}
  \vspace{-3mm}
\end{figure}

\paragraph{\textbf{Losses.}} The ablations of gradient loss and white balance loss are shown in Fig. \ref{fig:loss}. The model without gradient loss produces distorted colors due to an incorrect CRF in the green channel. The model without white balance loss produces results with random white balance, which is unacceptable. In contrast, the CRFs of the model with full loss terms are smooth and similar across all RGB channels. This matches the ground truth that each channel is tone-mapped with the same CRF in this synthetic scene. These ablations demonstrate that our loss terms encourage the camera model to correctly fit the global tone-mapping process.

\subsection{Limitations} 
Our method has a few limitations. Our method is trained per scene, which takes lots of time for optimization. However, the optimization time can't restrict our method to practical application, we can implement our model with the strategy of Instant NGP \cite{muller2022instant} that takes only 2.5 minutes to approximate an RGB image of resolution $20000 \times 23466$. In addition, the noise module is also an important component of the physical imaging formulation, especially when scenes are captured in low light. The noise model isn't simulated in our model, so the noise is recovered as part of an image. Finally, the results of our camera model are affected by the performance of the scene model. For example, the layered neural atlases model fails on dynamic scenes with complex geometry, self-occlusions, or extreme deformations with a single atlas layer (as shown in the supplementary). 

\section{Conclusion}
In this paper, we propose an interesting component for implicit neural representations, an implicit camera model, to simulate the physical imaging process. In particular, our camera model contains an implicit \textit{blur generator} module and an implicit \textit{tone mapper} module, to estimate the point spread function and camera response function respectively. It is jointly optimized with scene models to invert the imaging process under the supervision of visual signals with different focuses and exposures. To disentangle the camera imaging functions and combine various captured scenarios better, a set of regularization terms are introduced to leverage the geometry and camera knowledge to achieve image tasks, including HDR imaging and all-in-focus. Experiments on various tasks confirm the superiority of the proposed self-supervised implicit camera model. With simple modifications, the framework of our camera model can be adapted to solve other inverse imaging tasks. Our code and models will be released publicly to facilitate reproducible research.

\noindent\textbf{Acknowledgements.} 
The work was supported by NSFC under Grant 62031023. 

{\small
\bibliographystyle{ieee_fullname}
\bibliography{egbib}

\begin{thebibliography}{10}\itemsep=-1pt

\bibitem{barron2021mip}
Jonathan~T. Barron, Ben Mildenhall, Matthew Tancik, Peter Hedman, Ricardo
  Martin-Brualla, and Pratul~P. Srinivasan.
\newblock Mip-nerf: A multiscale representation for anti-aliasing neural
  radiance fields.
\newblock In {\em ICCV}, pages 5855--5864, October 2021.

\bibitem{blender}
Blender.
\newblock Blender project.
\newblock \url{https://www.blender.org/features/}, 2022.

\bibitem{boss2021nerd}
Mark Boss, Raphael Braun, Varun Jampani, Jonathan~T Barron, Ce Liu, and Hendrik
  Lensch.
\newblock {NeRD}: Neural reflectance decomposition from image collections.
\newblock In {\em ICCV}, pages 12684--12694, 2021.

\bibitem{campisi2017blind}
Patrizio Campisi and Karen Egiazarian.
\newblock {\em Blind image deconvolution: theory and applications}.
\newblock CRC press, 2017.

\bibitem{chen2021hdr}
Guanying Chen, Chaofeng Chen, Shi Guo, Zhetong Liang, Kwan-Yee~K Wong, and Lei
  Zhang.
\newblock Hdr video reconstruction: A coarse-to-fine network and a real-world
  benchmark dataset.
\newblock In {\em ICCV}, pages 2502--2511, 2021.

\bibitem{chen2021nerv}
Hao Chen, Bo He, Hanyu Wang, Yixuan Ren, Ser~Nam Lim, and Abhinav Shrivastava.
\newblock Nerv: Neural representations for videos.
\newblock {\em NeurIPS}, 34, 2021.

\bibitem{chen2012theoretical}
Xiaogang Chen, Feng Li, Jie Yang, and Jingyi Yu.
\newblock A theoretical analysis of camera response functions in image
  deblurring.
\newblock In {\em ECCV}, pages 333--346. Springer, 2012.

\bibitem{chen2021hallucinated}
Xingyu Chen, Qi Zhang, Xiaoyu Li, Yue Chen, Feng Ying, Xuan Wang, and Jue Wang.
\newblock Hallucinated neural radiance fields in the wild.
\newblock In {\em CVPR}, 2022.

\bibitem{debevec1997recovering}
Paul~E. Debevec and Jitendra Malik.
\newblock Recovering high dynamic range radiance maps from photographs.
\newblock In {\em SIGGRAPH}, page 369–378, 1997.

\bibitem{debevec2008recovering}
Paul~E Debevec and Jitendra Malik.
\newblock Recovering high dynamic range radiance maps from photographs.
\newblock In {\em ACM SIGGRAPH 2008 classes}, pages 1--10. 2008.

\bibitem{dupont2021coin}
Emilien Dupont, Adam Goli{\'n}ski, Milad Alizadeh, Yee~Whye Teh, and Arnaud
  Doucet.
\newblock Coin: Compression with implicit neural representations.
\newblock {\em arXiv preprint arXiv:2103.03123}, 2021.

\bibitem{grosch2006fast}
Thorsten Grosch et~al.
\newblock Fast and robust high dynamic range image generation with camera and
  object movement.
\newblock {\em Vision, Modeling and Visualization, RWTH Aachen}, 277284, 2006.

\bibitem{guo2019fusegan}
Xiaopeng Guo, Rencan Nie, Jinde Cao, Dongming Zhou, Liye Mei, and Kangjian He.
\newblock Fusegan: Learning to fuse multi-focus image via conditional
  generative adversarial network.
\newblock {\em IEEE TMM}, 21(8):1982--1996, 2019.

\bibitem{hu2018exposure}
Yuanming Hu, Hao He, Chenxi Xu, Baoyuan Wang, and Stephen Lin.
\newblock Exposure: A white-box photo post-processing framework.
\newblock {\em ACM TOG}, 37(2):1--17, 2018.

\bibitem{huang2021hdr}
Xin Huang, Qi Zhang, Feng Ying, Hongdong Li, Xuan Wang, and Qing Wang.
\newblock Hdr-nerf: High dynamic range neural radiance fields.
\newblock In {\em CVPR}, 2022.

\bibitem{jacobs2008automatic}
Katrien Jacobs, Celine Loscos, and Greg Ward.
\newblock Automatic high-dynamic range image generation for dynamic scenes.
\newblock {\em IEEE Computer Graphics and Applications}, 28(2):84--93, 2008.

\bibitem{kalantari2017deep}
Nima~Khademi Kalantari, Ravi Ramamoorthi, et~al.
\newblock Deep high dynamic range imaging of dynamic scenes.
\newblock {\em ACM TOG}, 36(4):144--1, 2017.

\bibitem{kaneko2022ar}
Takuhiro Kaneko.
\newblock Ar-nerf: Unsupervised learning of depth and defocus effects from
  natural images with aperture rendering neural radiance fields.
\newblock In {\em CVPR}, pages 18387--18397, 2022.

\bibitem{kasten2021layered}
Yoni Kasten, Dolev Ofri, Oliver Wang, and Tali Dekel.
\newblock Layered neural atlases for consistent video editing.
\newblock {\em ACM TOG}, 40(6):1--12, 2021.

\bibitem{li2019multi}
Huaguang Li, Rencan Nie, Jinde Cao, Xiaopeng Guo, Dongming Zhou, and Kangjian
  He.
\newblock Multi-focus image fusion using u-shaped networks with a hybrid
  objective.
\newblock {\em IEEE Sensors Journal}, 19(21):9755--9765, 2019.

\bibitem{li2013image}
Shutao Li, Xudong Kang, Jianwen Hu, and Bin Yang.
\newblock Image matting for fusion of multi-focus images in dynamic scenes.
\newblock {\em Information Fusion}, 14(2):147--162, 2013.

\bibitem{li2021neural}
Zhengqi Li, Simon Niklaus, Noah Snavely, and Oliver Wang.
\newblock Neural scene flow fields for space-time view synthesis of dynamic
  scenes.
\newblock In {\em CVPR}, pages 6498--6508, 2021.

\bibitem{liu2017multi}
Yu Liu, Xun Chen, Hu Peng, and Zengfu Wang.
\newblock Multi-focus image fusion with a deep convolutional neural network.
\newblock {\em Information Fusion}, 36:191--207, 2017.

\bibitem{ma2021deblur}
Li Ma, Xiaoyu Li, Jing Liao, Qi Zhang, Xuan Wang, Jue Wang, and Pedro~V Sander.
\newblock Deblur-nerf: Neural radiance fields from blurry images.
\newblock In {\em CVPR}, 2022.

\bibitem{mantiuk2011hdr}
Rafa{\l} Mantiuk, Kil~Joong Kim, Allan~G Rempel, and Wolfgang Heidrich.
\newblock Hdr-vdp-2: A calibrated visual metric for visibility and quality
  predictions in all luminance conditions.
\newblock {\em ACM TOG}, 30(4):1--14, 2011.

\bibitem{martin2021nerf}
Ricardo Martin-Brualla, Noha Radwan, Mehdi~SM Sajjadi, Jonathan~T Barron,
  Alexey Dosovitskiy, and Daniel Duckworth.
\newblock {NeRF} in the wild: Neural radiance fields for unconstrained photo
  collections.
\newblock In {\em CVPR}, pages 7210--7219, 2021.

\bibitem{mildenhall2021nerf}
Ben Mildenhall, Peter Hedman, Ricardo Martin-Brualla, Pratul Srinivasan, and
  Jonathan~T Barron.
\newblock Nerf in the dark: High dynamic range view synthesis from noisy raw
  images.
\newblock In {\em CVPR}, 2022.

\bibitem{mildenhall2020nerf}
Ben Mildenhall, Pratul~P Srinivasan, Matthew Tancik, Jonathan~T Barron, Ravi
  Ramamoorthi, and Ren Ng.
\newblock {NeRF}: Representing scenes as neural radiance fields for view
  synthesis.
\newblock In {\em ECCV}, pages 405--421. Springer, 2020.

\bibitem{muller2022instant}
Thomas M{\"u}ller, Alex Evans, Christoph Schied, and Alexander Keller.
\newblock Instant neural graphics primitives with a multiresolution hash
  encoding.
\newblock {\em arXiv preprint arXiv:2201.05989}, 2022.

\bibitem{nejati2015multi}
Mansour Nejati, Shadrokh Samavi, and Shahram Shirani.
\newblock Multi-focus image fusion using dictionary-based sparse
  representation.
\newblock {\em Information Fusion}, 25:72--84, 2015.

\bibitem{niu2021hdr}
Yuzhen Niu, Jianbin Wu, Wenxi Liu, Wenzhong Guo, and Rynson~WH Lau.
\newblock Hdr-gan: Hdr image reconstruction from multi-exposed ldr images with
  large motions.
\newblock {\em IEEE TIP}, 30:3885--3896, 2021.

\bibitem{ouyang2021neural}
Hao Ouyang, Zifan Shi, Chenyang Lei, Ka~Lung Law, and Qifeng Chen.
\newblock Neural camera simulators.
\newblock In {\em CVPR}, pages 7700--7709, 2021.

\bibitem{photomatix}
Photomatrix.
\newblock Photo editing software for hdr \& real estate photography.
\newblock \url{https://www.hdrsoft.com/}, 2021.

\bibitem{prabhakar2020towards}
K~Ram Prabhakar, Susmit Agrawal, Durgesh~Kumar Singh, Balraj Ashwath, and
  R~Venkatesh Babu.
\newblock Towards practical and efficient high-resolution {HDR} deghosting with
  {CNN}.
\newblock In {\em ECCV}, pages 497--513. Springer, 2020.

\bibitem{pumarola2021d}
Albert Pumarola, Enric Corona, Gerard Pons-Moll, and Francesc Moreno-Noguer.
\newblock D-nerf: Neural radiance fields for dynamic scenes.
\newblock In {\em CVPR}, pages 10318--10327, 2021.

\bibitem{sen2012robust}
Pradeep Sen, Nima~Khademi Kalantari, Maziar Yaesoubi, Soheil Darabi, Dan~B
  Goldman, and Eli Shechtman.
\newblock Robust patch-based hdr reconstruction of dynamic scenes.
\newblock {\em ACM TOG}, 31(6):203--1, 2012.

\bibitem{sitzmann2020implicit}
Vincent Sitzmann, Julien Martel, Alexander Bergman, David Lindell, and Gordon
  Wetzstein.
\newblock Implicit neural representations with periodic activation functions.
\newblock {\em NeurIPS}, 33:7462--7473, 2020.

\bibitem{sturge1989imaging}
J.M. Sturge, V. Walworth, and A. Shepp.
\newblock {\em Imaging Processes and Materials}.
\newblock Van Nostrand Reinhold, 1989.

\bibitem{su2017deep}
Shuochen Su, Mauricio Delbracio, Jue Wang, Guillermo Sapiro, Wolfgang Heidrich,
  and Oliver Wang.
\newblock Deep video deblurring for hand-held cameras.
\newblock In {\em CVPR}, pages 1279--1288, 2017.

\bibitem{tang2018pixel}
Han Tang, Bin Xiao, Weisheng Li, and Guoyin Wang.
\newblock Pixel convolutional neural network for multi-focus image fusion.
\newblock {\em Information Sciences}, 433:125--141, 2018.

\bibitem{teed2020raft}
Zachary Teed and Jia Deng.
\newblock Raft: Recurrent all-pairs field transforms for optical flow.
\newblock In {\em ECCV}, pages 402--419. Springer, 2020.

\bibitem{tursun2015state}
Okan~Tarhan Tursun, Ahmet~O{\u{g}}uz Aky{\"u}z, Aykut Erdem, and Erkut Erdem.
\newblock The state of the art in {HDR} deghosting: a survey and evaluation.
\newblock In {\em Comput. Graph. Forum}, volume~34, pages 683--707. Wiley
  Online Library, 2015.

\bibitem{wang2021deep}
Chengyu Wang, Qian Huang, Ming Cheng, Zhan Ma, and David~J Brady.
\newblock Deep learning for camera autofocus.
\newblock {\em IEEE Transactions on Computational Imaging}, 7:258--271, 2021.

\bibitem{wang2020novel}
Chang Wang, Zongya Zhao, Qiongqiong Ren, Yongtao Xu, and Yi Yu.
\newblock A novel multi-focus image fusion by combining simplified very deep
  convolutional networks and patch-based sequential reconstruction strategy.
\newblock {\em Applied Soft Computing}, 91:106253, 2020.

\bibitem{wu2018deep}
Shangzhe Wu, Jiarui Xu, Yu-Wing Tai, and Chi-Keung Tang.
\newblock Deep high dynamic range imaging with large foreground motions.
\newblock In {\em ECCV}, pages 117--132, 2018.

\bibitem{wu2022dof}
Zijin Wu, Xingyi Li, Juewen Peng, Hao Lu, Zhiguo Cao, and Weicai Zhong.
\newblock Dof-nerf: Depth-of-field meets neural radiance fields.
\newblock In {\em ACM MM}, pages 1718--1729, 2022.

\bibitem{xu2020u2fusion}
Han Xu, Jiayi Ma, Junjun Jiang, Xiaojie Guo, and Haibin Ling.
\newblock U2fusion: A unified unsupervised image fusion network.
\newblock {\em IEEE TPAMI}, 44(1):502--518, 2020.

\bibitem{xu2020fusiondn}
Han Xu, Jiayi Ma, Zhuliang Le, Junjun Jiang, and Xiaojie Guo.
\newblock Fusiondn: A unified densely connected network for image fusion.
\newblock In {\em Proceedings of the AAAI Conference on Artificial
  Intelligence}, volume~34, pages 12484--12491, 2020.

\bibitem{yan2019attention}
Qingsen Yan, Dong Gong, Qinfeng Shi, Anton van~den Hengel, Chunhua Shen, Ian
  Reid, and Yanning Zhang.
\newblock Attention-guided network for ghost-free high dynamic range imaging.
\newblock In {\em CVPR}, pages 1751--1760, 2019.

\bibitem{yan2020deep}
Qingsen Yan, Lei Zhang, Yu Liu, Yu Zhu, Jinqiu Sun, Qinfeng Shi, and Yanning
  Zhang.
\newblock Deep hdr imaging via a non-local network.
\newblock {\em IEEE TIP}, 29:4308--4322, 2020.

\bibitem{yan2019robust}
Qingsen Yan, Yu Zhu, and Yanning Zhang.
\newblock Robust artifact-free high dynamic range imaging of dynamic scenes.
\newblock {\em Multimedia Tools and Applications}, 78(9):11487--11505, 2019.

\bibitem{yang2019multilevel}
Yong Yang, Zhipeng Nie, Shuying Huang, Pan Lin, and Jiahua Wu.
\newblock Multilevel features convolutional neural network for multifocus image
  fusion.
\newblock {\em IEEE Transactions on Computational Imaging}, 5(2):262--273,
  2019.

\bibitem{yin2021virtual}
Wei Yin, Yifan Liu, and Chunhua Shen.
\newblock Virtual normal: Enforcing geometric constraints for accurate and
  robust depth prediction.
\newblock {\em IEEE TPAMI}, 2021.

\bibitem{yu2021pixelnerf}
Alex Yu, Vickie Ye, Matthew Tancik, and Angjoo Kanazawa.
\newblock {pixelNeRF}: Neural radiance fields from one or few images.
\newblock In {\em CVPR}, pages 4578--4587, 2021.

\bibitem{yu2021reconfigisp}
Ke Yu, Zexian Li, Yue Peng, Chen~Change Loy, and Jinwei Gu.
\newblock Reconfigisp: Reconfigurable camera image processing pipeline.
\newblock In {\em ICCV}, pages 4248--4257, 2021.

\bibitem{zhang2021mff}
Hao Zhang, Zhuliang Le, Zhenfeng Shao, Han Xu, and Jiayi Ma.
\newblock Mff-gan: An unsupervised generative adversarial network with adaptive
  and gradient joint constraints for multi-focus image fusion.
\newblock {\em Information Fusion}, 66:40--53, 2021.

\bibitem{zhang2018unreasonable}
Richard Zhang, Phillip Isola, Alexei~A Efros, Eli Shechtman, and Oliver Wang.
\newblock The unreasonable effectiveness of deep features as a perceptual
  metric.
\newblock In {\em CVPR}, pages 586--595, 2018.

\bibitem{zhang2021deep}
Xingchen Zhang.
\newblock Deep learning-based multi-focus image fusion: A survey and a
  comparative study.
\newblock {\em IEEE TPAMI}, 2021.

\bibitem{zhang2020ifcnn}
Yu Zhang, Yu Liu, Peng Sun, Han Yan, Xiaolin Zhao, and Li Zhang.
\newblock Ifcnn: A general image fusion framework based on convolutional neural
  network.
\newblock {\em Information Fusion}, 54:99--118, 2020.

\end{thebibliography}
}

\clearpage
\section*{Supplemental Materials}
\setcounter{section}{0}
\renewcommand\thesection{\Alph{section}}

\section{Additional Implementation Details}

\subsection{Network Details}
All networks in our framework are based on the MLP architecture. The deformation network $\mathcal{D}$ is a 4-layer MLP with 256 channels of each layer. The atlas network $\mathcal{A}$ consists of 4 layers with 512 channels. The offset network $\mathcal{O}$ and weight network $\mathcal{W}$ also have 4 layers, each with 64 channels. The tone-mapping network $\mathcal{T}$ is composed of three MLPs, each of 2 layers with 128 channels, to fit the response functions of R, G, and B channels respectively. Rectified Linear Unit (ReLU) activations are adopted between inner layers of networks, and the outputs of the last layers are passed through a tanh activation, except for the weight network $\mathcal{W}$ which takes softmax activation instead.

\subsection{Training Details}

Without the supervision of all-in-focus HDR images, our framework is sensitive to the initial values of parameters of the network. During the initial bootstrapping phase ($10$k iterations), we firstly train the deformation network $\mathcal{D}$ by mapping pixel position $\mathbf{p}=(x,y,i)$ (normalized to range $[-1,1]$) to coordinate $(x,y)$. It enforces the deformations to be initialized as zero, considering that the static background occupies a large proportion of the image sequence. 
The optical flow estimated by the off-the-shelf method \cite{teed2020raft} is not always accurate due to the different focuses and exposures of input images. Therefore, during the training phase, the flow loss weight $\lambda_f$ gradually decays to $0$ over the course of optimization. 


\section{Additional Experiments and Results}
\subsection{Comparisons with Traditional Methods}
\begin{figure*}[!t]
    \vspace{-8mm}
    \centering
    \includegraphics[width=\linewidth]{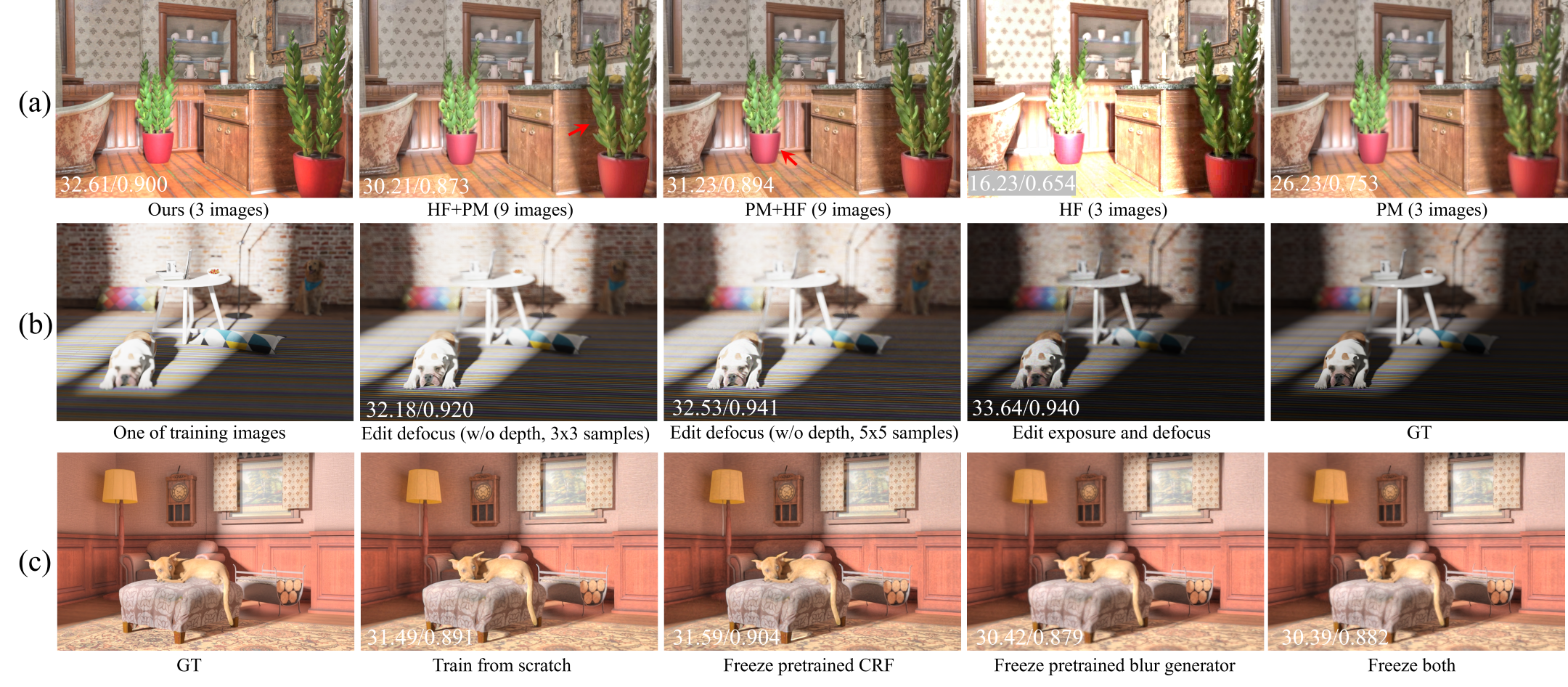}
    \vspace{-8mm}
    \caption{(a) Comparisons with two off-the-shelf optimization-based algorithms. ``HF" is Helicon Focus software for all-in-focus images and ``PM" is Photomatix software for HDR images. (b) Evaluations of our generated LDR and defocused images. (c) Evaluations on a new scene by freezing the pre-trained camera model. The first and third rows are all-in-focus HDR images. The PSNR and SSIM are presented in the lower left corner. \textbf{Please open with PDF reader for zoom-in}. }
    \label{fig:rebuttal}
    \vspace{-2mm}
\end{figure*}

Compared with optimization-based traditional methods on the all-in-focus HDR imaging task, our neural camera model enables recovering irradiance maps from the image stack where exposure and defocused blur vary simultaneously. As shown in \cref{fig:rebuttal} (a), HF fails to recover an all-in-focus image due to the different exposures of input images, and PM is similar. Although HF$+$PM and PM$+$HF can recover all-in-focus HDR images from 9 images, our method takes only 3 images as input and outperforms them. 

\begin{figure}[!t]
  \centering
  \includegraphics[width=\linewidth]{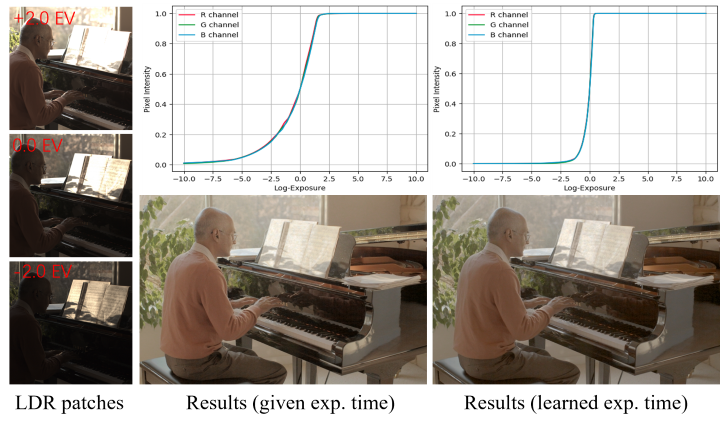}
  \caption{Results comparison of our method with given exposure time and learned exposure time on an ME scene. The left column shows the zoom-in patches of input images. The top row shows CRFs learned by \textit{tone mapper}. The bottom row shows our tone-mapped HDR results.}
  \label{fig:learnexp}
\end{figure}

\subsection{Larger Sampling Patch}
The evaluation of our generated LDR and defocused images are presented in \cref{fig:rebuttal} (b). Using depth maps is helpful to generate accurate defocused blur. However, our method can also produce photorealistic defocused blur without the monocular depth. Using $3\times3$ samples to represent the PSF is not enough when the degree of blur is too large, which causes aliasing artifacts.  The blur pattern of defocus is more reasonable when we use $5\times5$ pixels to infer a defocused pixel, as shown in \cref{fig:rebuttal} (b). 

\subsection{Evaluation of Implicit Camera Model}
We evaluate the pre-trained camera model on a new scene. The results are shown in \cref{fig:rebuttal} (c). One can see that our model achieves a competitive result using a pretrained CRF.  Unlike our \textit{tone mapper}, pixel positions are fed into our \textit{blur generator} to produce blending weights and offsets, which causes the learned blur generator to depend on the trained scene. Therefore, the performance decreases when we freeze the pre-trained blur generator and then train our model on a new scene.

\subsection{Learned Exposure}
The EXIF tags may be unavailable for compressed images from the internet. So we can also learn the exposure time for each image during the optimization. Figure \ref{fig:learnexp} shows the recovered HDR images with given exposure time or learned exposure time on the ME dataset. As can be seen, there is a scale difference between the two CRFs but HDR images with abundant details are well reconstructed by the two models, which illustrates learning exposure time is feasible. 

\begin{figure}[!t]
  \centering
  \includegraphics[width=\linewidth]{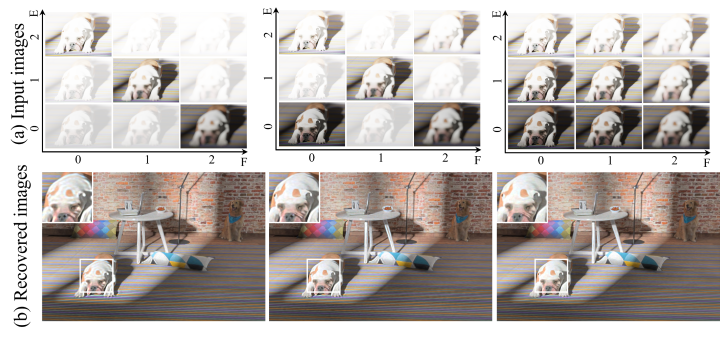}
  \caption{Results comparison of our method with different sets of input images on a synthetic MFME scene. (a) Top row shows the zoom-in patches of the input images. $F$ denotes the focus and $E$ denotes the exposure. (b) Bottom row shows the corresponding tone-mapped results. \bf{Better viewed on screen with zoom in.}}
  \label{fig:traingmode}
\end{figure}

\subsection{The Number of Input (Training) images.} To evaluate the influence of input images, different combinations of exposures and focuses are evaluated in our method. Figure \ref{fig:traingmode} shows three sets (3 images, 5 images, and 9 images) of input images. Ideally, 3 images are enough for our method to recover all-in-focus HDR images. However, in some special cases when images focus on over-exposed or under-exposed areas, the method produces results with artifacts (\textit{e.g.} artifacts on the head of the dog when the model is trained on 3 images). In these cases, raising the number of input images to 5 or 9 yields better results.

\subsection{More Results}

\begin{figure*}[!t]
  \centering
  \includegraphics[width=\textwidth]{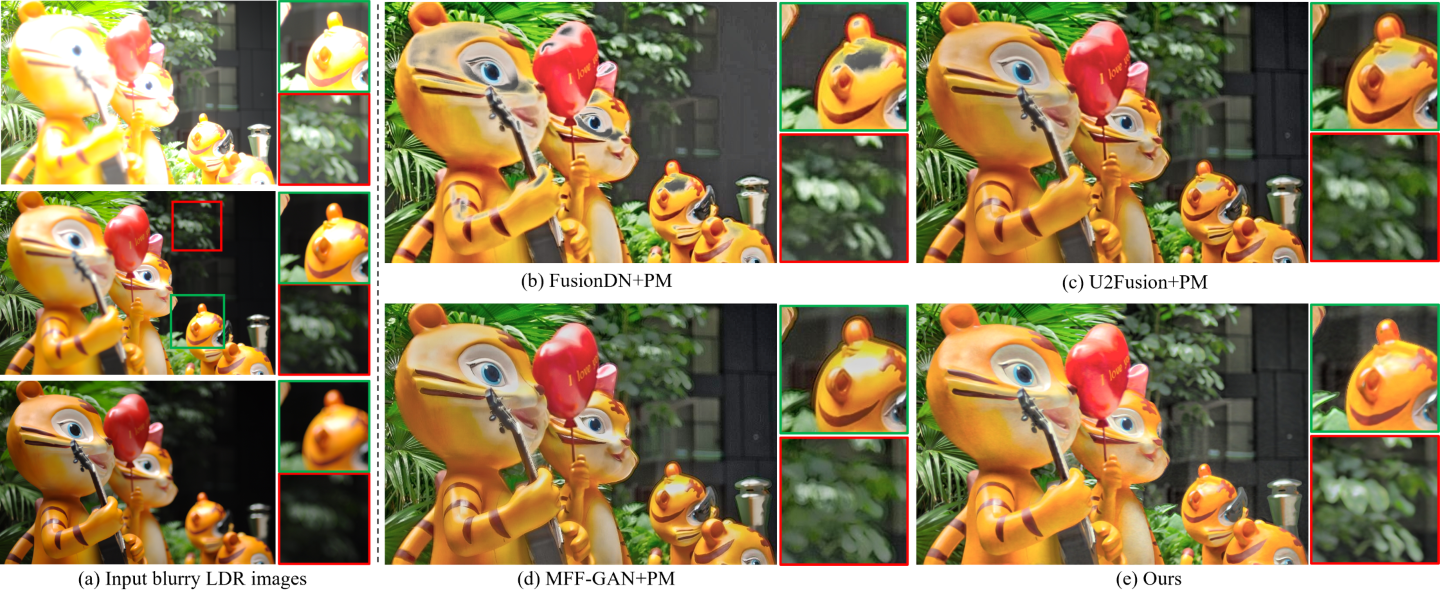}
  \caption{Example results of our method compared with two-stage methods on the MFME real dataset.  ``PM'' denotes the HDR imaging method in Photomatix \cite{photomatix}. (a) Our input images with different focuses and exposures. (b-e) All-in-focus and HDR images produced by three two-stage methods and our method. The red and green insets show the zoom-in views of the images. All HDR images are tone-mapped for display. }
  \label{fig:deblur_HDR_real}
\end{figure*}

\begin{figure*}[!b]
  \centering
  \includegraphics[width=\textwidth]{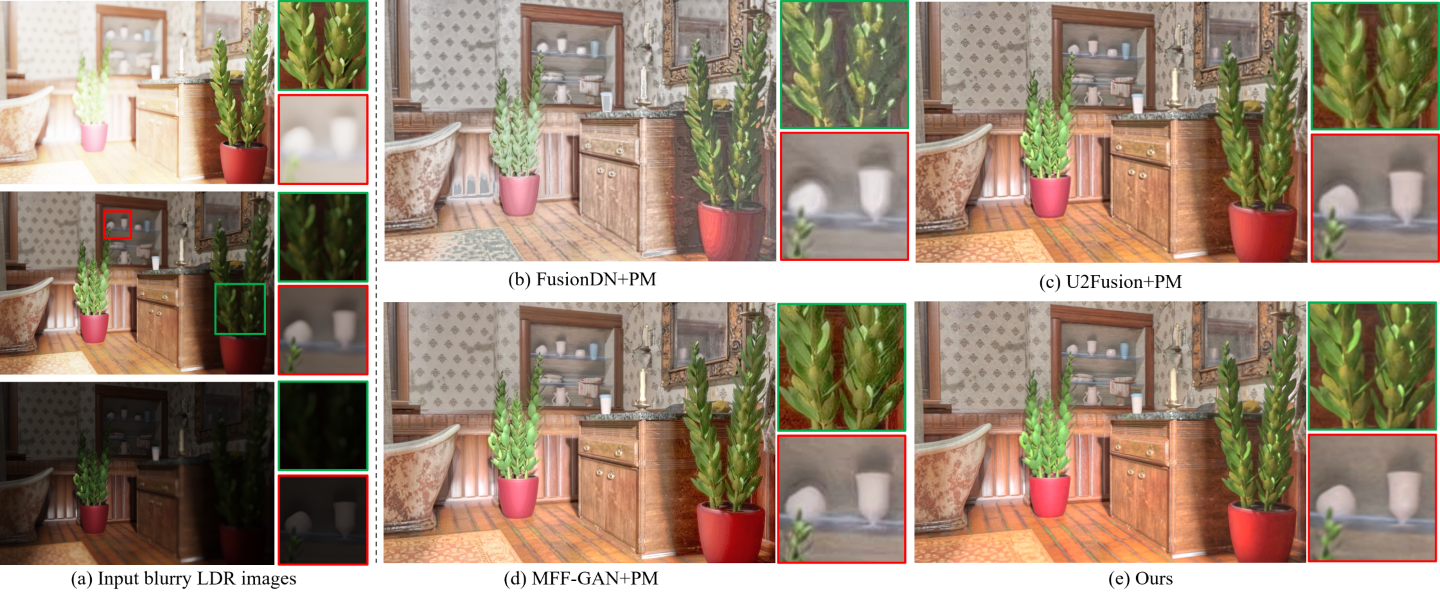}
  \caption{Example results of our method compared with two-stage methods on the MFME synthetic dataset.  ``PM'' denotes the HDR imaging method in Photomatix \cite{photomatix}. (a) Our input images with different focuses and exposures. (b-e) All-in-focus and HDR images produced by three two-stage methods and our method. The red and green insets show the zoom-in views of the images. All HDR images are tone-mapped for display.}
  \label{fig:supp_deblur_HDR_syn}
\end{figure*}

In Fig. \ref{fig:deblur_HDR_real} and \ref{fig:supp_deblur_HDR_syn}, we show the comparisons of our method with the two-stages methods for recovering all-in-focus and HDR images from the images with different focuses and exposures. As one can see, The results by FusionDN \cite{xu2020fusiondn} + PM \cite{photomatix} have distorted colors. U2Fuison \cite{xu2020u2fusion} + PM\cite{photomatix} and MFF-GAN \cite{zhang2021mff} + PM\cite{photomatix} produce better results with consistent colors, but both methods fail to deal with the ghosting near the object boundary, such as the cups in Fig. \ref{fig:supp_deblur_HDR_syn} (green insets). Compared with the two-stage methods, our method produces all-in-focus and HDR images with sharp boundaries and details. 

\begin{figure*}[!tb]
    \centering
    \includegraphics[width=\textwidth]{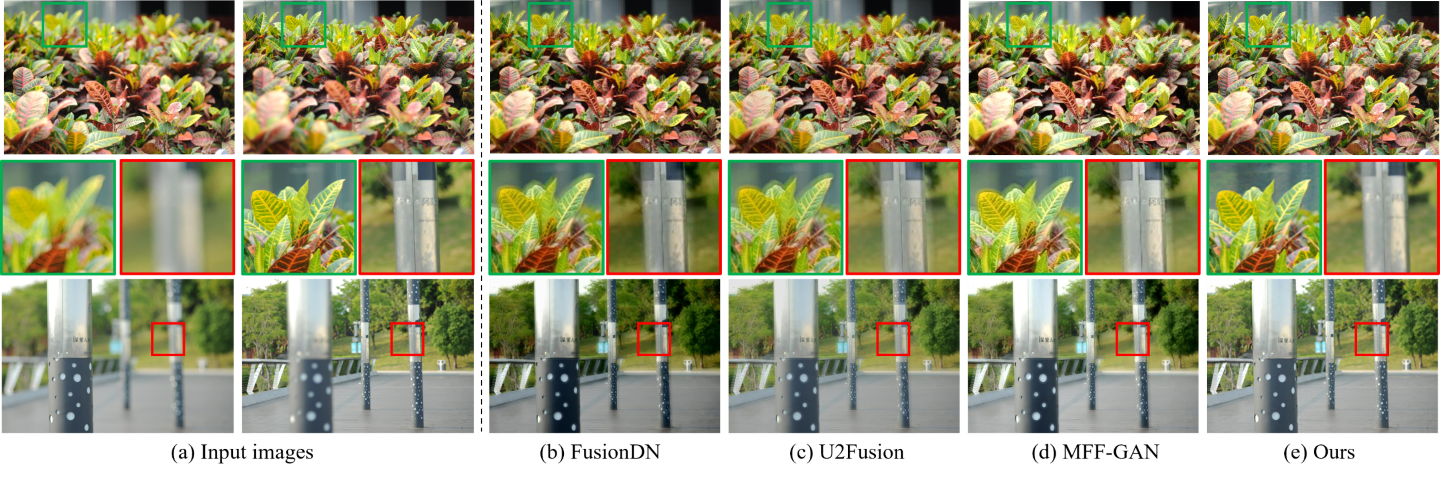}
    \caption{Example results of our method compared with MFIF methods on the MF dataset. (a) Two input  images. One is near-focused and the other is far-focused. (b-e) All-in-focus results by MFIF methods and our method. The red and green insets show the zoom-in views of the images. \bf{Better viewed on screen with zoom in.}}
    \label{fig:supp_deblur}
\end{figure*}

\begin{figure*}[h]
  \centering
  \includegraphics[width=\textwidth]{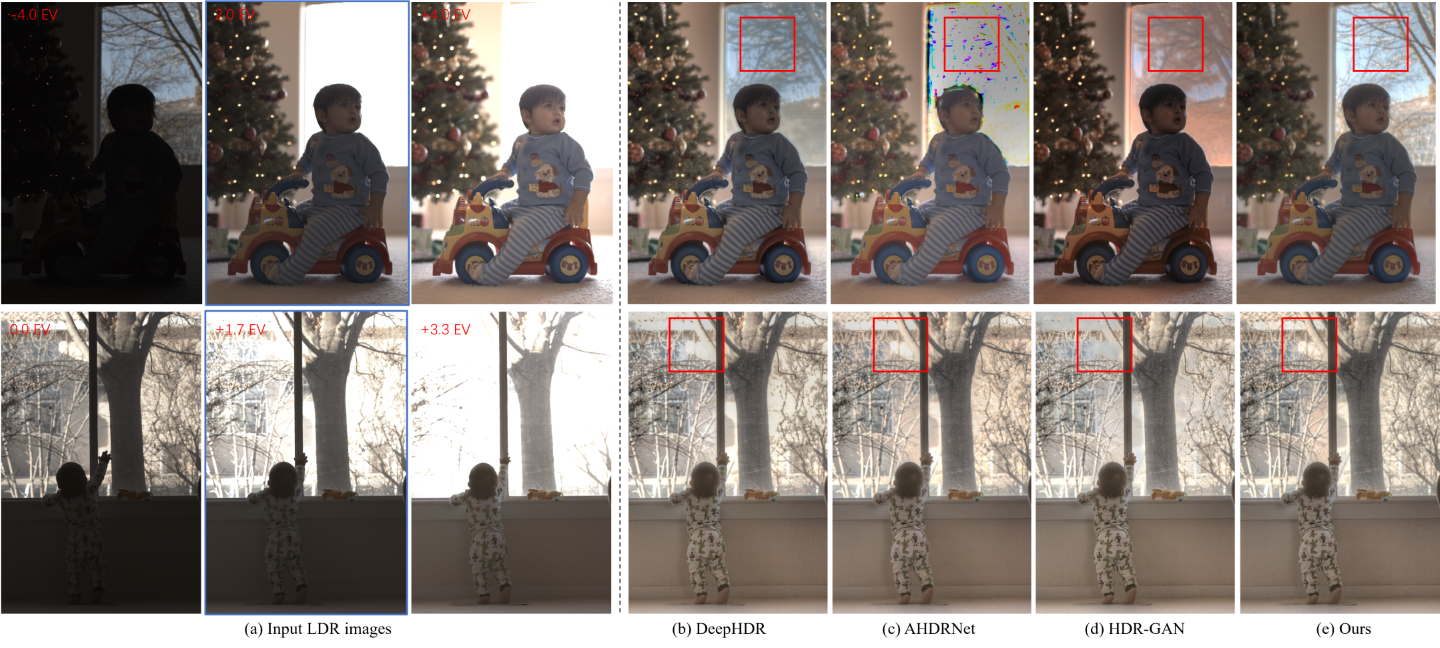}
  \caption{Example results of our method compared with HDR imaging methods on the ME dataset. (a) Three input images with different exposures. Exposure values (EVs) are shown in the upper left. The image highlight with a blue box denotes the reference image. (b-e) The recovered HDR images for the reference image. All HDR images are tone-mapped for display.}
  \label{fig:supp_HDR}
\end{figure*}

In Fig. \ref{fig:supp_deblur}, we show the comparisons of our method with multi-focus image fusion (MFIF) methods for recovering all-in-focus images from a near-focused image and a far-focused image. Similarly, the results by FusionDN \cite{xu2020fusiondn}, U2Fuison \cite{xu2020u2fusion} and MFF-GAN \cite{zhang2021mff} all have ghosting near boundaries, while our results are clearer and have a consistent color with input images.  Figure \ref{fig:supp_HDR} presents the recovered HDR results of our method and the state-of-the-art  HDR imaging methods (HDR-GAN \cite{niu2021hdr}, AHDRNet \cite{yan2019attention}, and DeepHDR \cite{wu2018deep}) for dynamic scenes. Three SOTA methods fail to recover the textures outside the window in the top scene, due to there the large over-exposed region in the reference image. Compared with them, our methods produce superior results. Besides, our method does not produce artifacts on the moving objects, such as the hand of the baby in the bottom scene, which demonstrates that our method can fit the scene with mall motions.

\begin{figure}[!t]
  \centering
  \includegraphics[width=\linewidth]{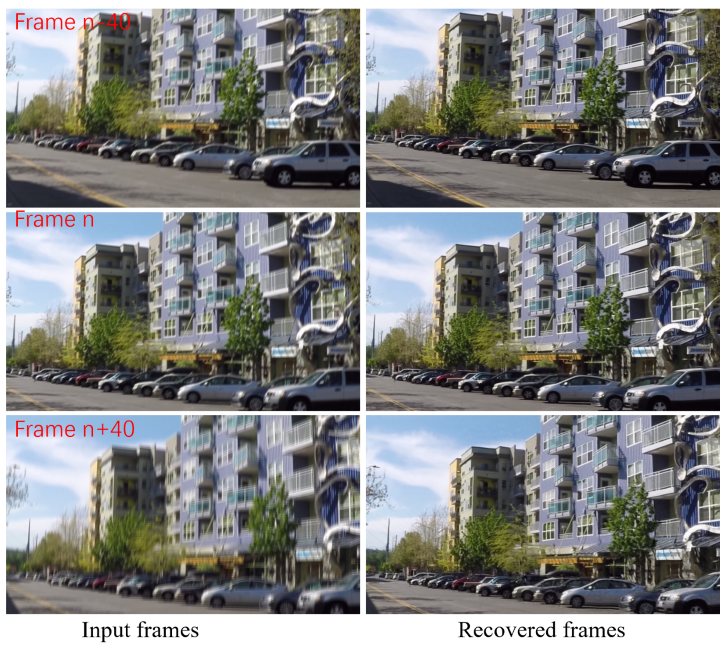}
  \vspace{-5mm}
  \caption{Visualization of our results for video deblurring. }
  \label{fig:videodebulur}
\end{figure}

\section{Applications}
\subsection{Controllable Rendering} 
\begin{figure*}[!t]
  \centering
  \includegraphics[width=\textwidth]{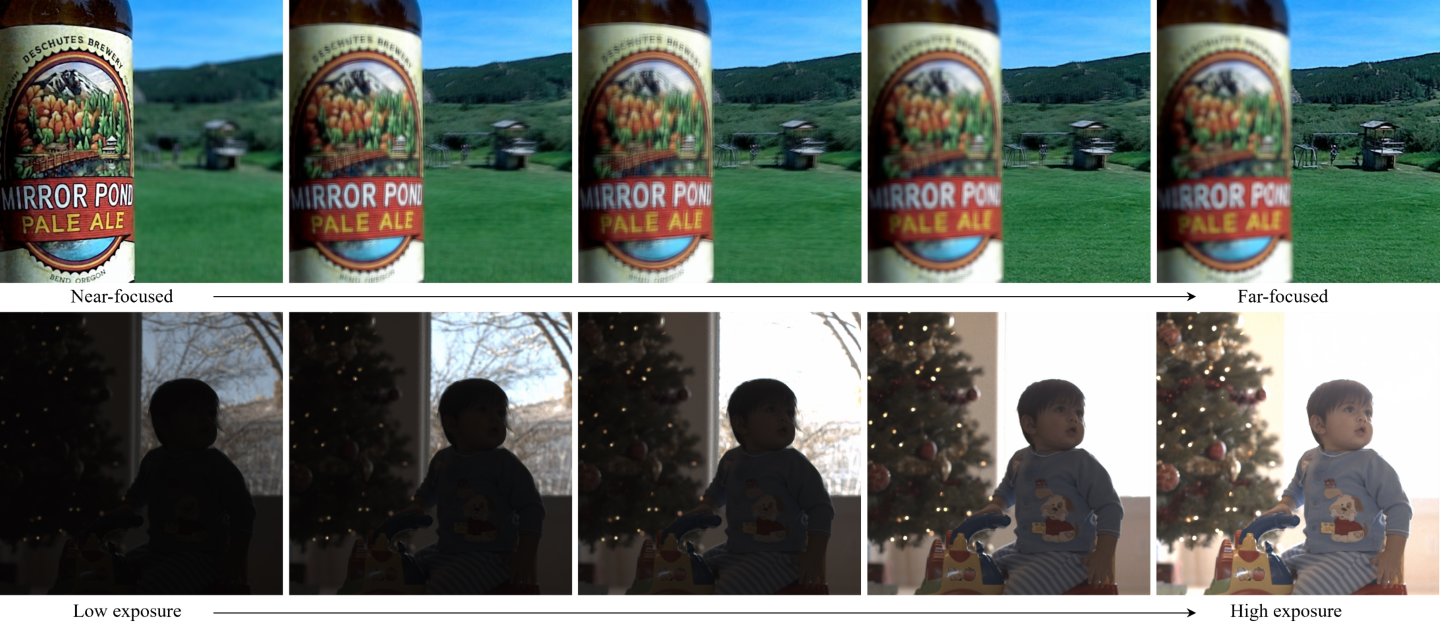}
  \caption{Controllable rendering results of our method. The leftmost and rightmost images are two training images, and the middle results are rendered with interpolated focus or exposure.}
  \label{fig:control}
\end{figure*}

The other consequence of our implicit camera model is that it enables rendering images with modified camera settings. When we keep the \textit{blur generator} and \textit{tone mapper} during the inference, our method can control the focus and exposure of rendered images. The degree of defocus blur is greatly related to depth. For example, the points at the same depth should have a consistent blur on images. To control the focus correctly, we concatenate the position $\mathbf{p}$ with the corresponding depth and feed them into the \textit{blur generator} to learn the PSF, where the depth is estimated using a single image depth estimation method \cite{yin2021virtual}. We have tried to modify the focus of the images from our MFME dataset, but we find the depth estimation method failed to predict accurate depths. Consequently, we evaluate the focus control on a Lytro dataset \cite{nejati2015multi}. The scene contents in the Lytro dataset are relatively simple, so we can estimate the depth accurately. To render images with varying focus, we interpolate the image indices $i$ that are fed into the \textit{blur generator}. Additionally, we control the exposure of rendered images by modifying the exposure time $\Delta t$. The exposure control is evaluated on the ME dataset. Figure \ref{fig:control} shows the controllable rendering of our method. We see the focus of the images (top row) smoothly varies from the foreground to the background. The bottom row presents the modification of exposures, where the exposure of the renderings increases gradually.


\begin{figure}[!t]
  \centering
  \includegraphics[width=\linewidth]{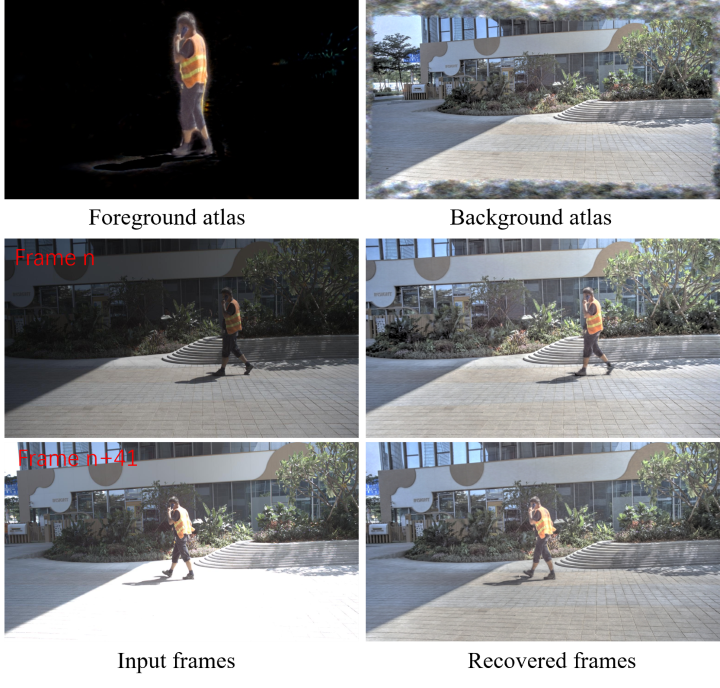}
  \vspace{-5mm}
  \caption{Visualization of our results for HDR video reconstruction. The visualizations of the neural atlases are shown in the first row.}
  \label{fig:videohdr}
  \vspace{-5mm}
\end{figure}

\subsection{Video Enhancement} \label{sec:videodata}
Our implicit camera model is also applicable to video enhancement combined with video scene representations. We adopt the layered neural atlases representation \cite{kasten2021layered}, which decomposes the video into a set of layered 2D atlases to deal with object motions and camera motions. We evaluate our model for video deblurring on Deep Video Deblurring (DVD) dataset \cite{su2017deep} and HDR video reconstruction on the Deep HDR Video (DHV) dataset \cite{chen2021hdr}. A video deblurring case is shown in Fig. \ref{fig:videodebulur}. The input video of $100$ frames with camera motion blur and our method recovers sharper textures. For the HDR video reconstruction task, the input is a video of $80$ frames with alternating exposures. Note that, this input video contains a moving person, so the video is represented with two atlases: an atlas for the foreground and an atlas for the background. In Fig. \ref{fig:videohdr}, we show the results for HDR video reconstruction. We can see that the scene contents are successfully split into two atlases and our method recovers the texture of over-exposed areas based on information from other frames with a lower exposure (see the ground in frame $n+41$).

\subsection{A Failure Case}
\begin{figure}[!h]
  \centering
  \includegraphics[width=\linewidth]{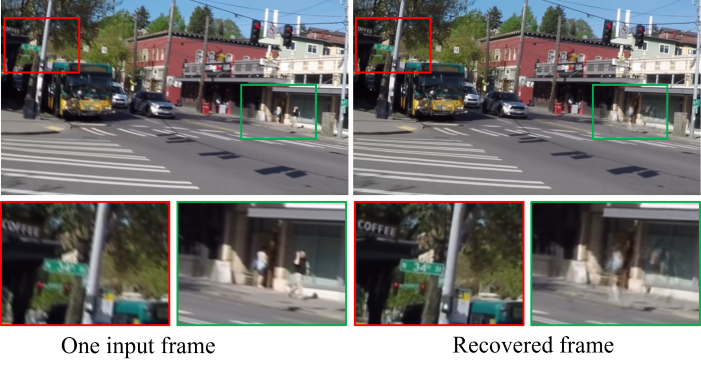}
  \vspace{-5mm}
  \caption{One failure case of our method. The input video is challenging in that the pedestrians have complex self-occlusions. The pedestrians on the street are missing in our recovered frames (see the green insets), while our method removes the camera motion blur of the video (see the red insets). }
  \label{fig:failed_case}
  \vspace{-5mm}
\end{figure}

Figure \ref{fig:failed_case} shows a failure case where pedestrians on the street are missing in the recovered images since the people are too small to split into a single atlas and there are lots of self-occlusions. However, our camera model also successfully removes the camera motion blur of the video.

\end{document}